\pdfoutput=1

\documentclass[11pt]{article}

\usepackage[]{emnlp2021}

\usepackage{times}
\usepackage{latexsym}

\usepackage{multirow}
\usepackage{mathrsfs}
\usepackage{amsfonts,amssymb}
\usepackage{amsmath}
\usepackage{booktabs}
\usepackage{graphicx}
\usepackage{subfigure}
\usepackage{diagbox}
\usepackage[T1]{fontenc}

\usepackage[utf8]{inputenc}

\usepackage{microtype}

\title{Detecting Speaker Personas from Conversational Texts}

\author{Jia-Chen Gu$^1$, Zhen-Hua Ling$^1$\thanks{\hspace{1.5mm}Corresponding author.}, Yu Wu$^2$, Quan Liu$^{1,3}$, Zhigang Chen$^3$, Xiaodan Zhu$^4$ \\
  $^1$National Engineering Laboratory for Speech and Language Information Processing, \\
      University of Science and Technology of China, Hefei, China \\
  $^2$Microsoft Research Asia, Beijing, China \\
  $^3$State Key Laboratory of Cognitive Intelligence, iFLYTEK Research, Hefei, China \\
  $^4$ECE \& Ingenuity Labs, Queen's University, Kingston, Canada \\
{\tt gujc@mail.ustc.edu.cn}, {\tt \{zhling, quanliu\}@ustc.edu.cn}, \\ 
{\tt Wu.Yu@microsoft.com}, {\tt zgchen@iflytek.com}, {\tt xiaodan.zhu@queensu.ca}
}

\begin{document}
\maketitle
\begin{abstract}
  Personas are useful for dialogue response prediction.
  However, the personas used in current studies are pre-defined and hard to obtain before a conversation.
  To tackle this issue, we study a new task, named Speaker Persona Detection (SPD), which aims to detect speaker personas based on the plain conversational text.
  In this task, a best-matched persona is searched out from candidates given the conversational text.
  This is a \emph{many-to-many} semantic matching task because both contexts and personas in SPD are composed of multiple sentences.
  The long-term dependency and the dynamic redundancy among these sentences increase the difficulty of this task.
  We build a dataset for SPD, dubbed as \texttt{Persona Match on Persona-Chat (PMPC)}. 
  Furthermore, we evaluate several baseline models and propose utterance-to-profile (U2P) matching networks for this task.
  The U2P models operate at a fine granularity which treat both contexts and personas as sets of multiple sequences.
  Then, each sequence pair is scored and an interpretable overall score is obtained for a context-persona pair through aggregation.
  Evaluation results show that the U2P models outperform their baseline counterparts significantly.
\end{abstract}

\section{Introduction}

  \begin{figure}[t]
    \centering
    \includegraphics[width=7.5cm]{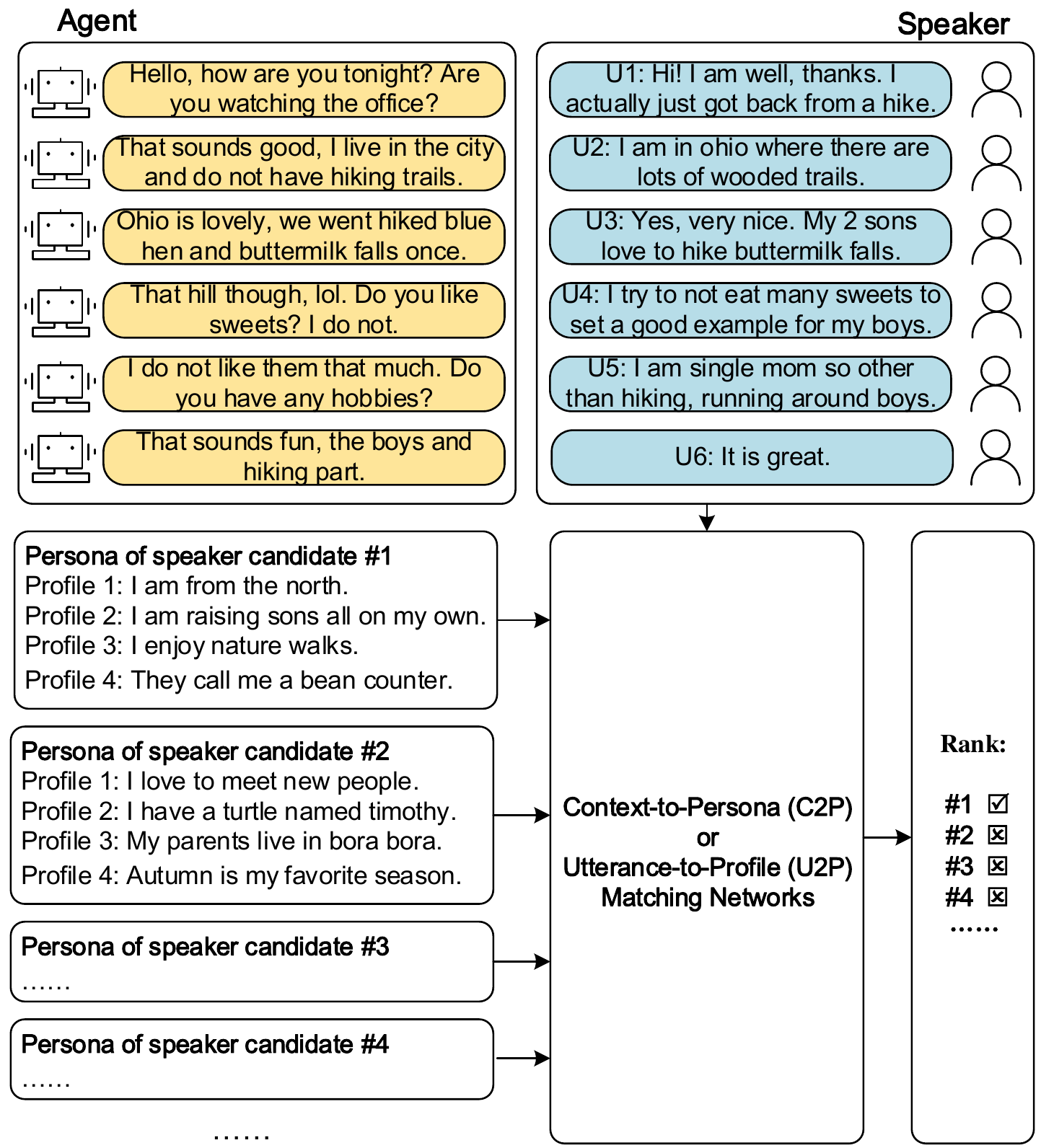}
    \caption{Illustration of our proposed SPD task. The matching network judges whether a persona candidate matches with the conversational texts of the speaker.}
    \label{fig4}
  \end{figure}

  Recently, human-machine conversation has achieved great success~\cite{lowe2015ubuntu,serban2016building,wu2017sequential,madotto2018mem2seq,tao2019one,zhang2020dialogpt,roller2021recipes}, and has been applied to various scenarios, such as customer service and conversational recommendation engine.
  It is well-known that a user's persona can help the machine to generate more appropriate responses.
  Many studies investigate how to predict a response if a persona is given.
  \citet{zhang2018personalizing,liu2020you,song2021bob} built dialogue agents to perceive the user's persona and then generated personalized responses.
  \citet{mazare2018training,gu2019dually,hua2020learning,gu2020filtering,zhu2021content} built matching networks to select a response matching not only the conversational context, but also the user's persona.

  However, these personas are pre-defined and difficult to obtain before a conversation, because the speaker might not want to fill out a specific table to show its persona due to privacy issues.
  Hence, the cold-start problem may hinder the persona-aware response prediction in practice.
  To tackle this issue, our intuition is that the personal information such as hobbies or occupations may be mentioned explicitly or implicitly during a conversation, which can be utilized to identify the speaker's persona.
  If we can get the persona from early conversations, it can be utilized for future persona-aware response prediction.

  Motivated by this, we propose a new task, named Speaker Persona Detection (SPD), which aims to detect a speaker's persona from the plain conversational text.
  As illustrated in Figure~\ref{fig4}, the agent proactively leads the conversation with a goal to collect the speaker information implicitly.
  Through the conversation between the speaker and the agent, we can use the speaker's utterances carrying the personal information to search for a best-matched persona in the database.
  Here, a persona description is composed of several profiles characterizing a person, which is unstructured and common in practice \cite{zhang2018personalizing,mazare2018training,gu2019dually,liu2020you,gu2021partner}.
  Then, the searched persona is utilized for follow-up personalized response generation or selection.
  The ability to learn speakers' personas can have wide applications in commercial chatbots, recommendation systems and other scenarios that involve conversations.
  For example, it can be applied to the recommendation system.
  Speakers first talk to the intelligent agent in their smart devices for several turns.
  Then the conversational text of the speaker is collected and utilized as a query to search for a best-matched persona description from candidates recorded in the database.
  Finally, the user preferences linked to the best-matched persona will be recommended to the speaker for providing personalized service.

  In our proposed SPD task, a conversation context is composed of multiple utterances and a persona is composed of multiple profiles, which brings three challenges to existing studies on matching.
  First, the matching in SPD relies on modeling the long-term dependency among conversation utterances.
  Second, the matching in SPD is established between two sets of sentences which requires a complicated \emph{many-to-many} matching framework.
  This has not been explored yet, as previous studies have been conducted either between a pair of sentences (i.e., \emph{one-to-one}) \cite{wang2013dataset,bowman2015large,williams2018broad}, or between a set of sentences and a single sentence (i.e., \emph{many-to-one}) \cite{lowe2015ubuntu,lai2017natural,wu2017sequential}.
  Third, there exists dynamic redundancy among conversation utterances and persona profiles.
  Specifically, the informativeness of each conversation utterance changes when inferring different profiles in a persona.
  Besides, some utterances may carry no personal information and some profiles may not be reflected by any utterances in the conversation.

  A dataset is built for studying the SPD task.
  We transform the existing \texttt{Persona-Chat} dataset \cite{zhang2018personalizing} by assuming that conversation sessions are given while personas are what models should predict.
  We name the transformed dataset \texttt{Persona Match on Persona-Chat (PMPC)}. 
  Regarding with SPD methods, context-to-persona (C2P) matching networks, which are established at a coarse granularity by concatenating two sets of sentences respectively, are employed as baselines.
  Furthermore, we propose utterance-to-profile (U2P) matching networks which are established at a fine granularity by treating each sentence in either contexts or personas individually.
  They first obtain the representation for each sentence and then derive the representations of contexts and personas through an aggregation operation.
  In this paper, (1) sentence-encoding-based framework such as the encoder of bag-of-words (BOW) \cite{joulin2017bag}, BiLSTM \cite{hochreiter1997long} and Transformer \cite{vaswani2017attention}, and (2) cross-attention-based framework such as ESIM \cite{chen2017enhanced} and (3) pretraining-based framework such as BERT \cite{devlin2019bert} are considered when building either C2P and U2P models.
  The experimental results comparing the C2P and U2P models demonstrate the effectiveness of the latter for solving the SPD task, which relies on the \emph{many-to-many} matching between two sets of sentences.
  In addition to the performance improvement, U2P models yield interpretability by explicitly scoring each utterance-profile pair and performing the aggregation operation. 

  In summary, our contributions in this paper are three-fold.
  (1) We propose a new task, Speaker Persona Detection (SPD), and construct a dataset for studying this problem, which make the first attempt to detect speaker personas from conversational texts by persona matching.
  (2) Many baseline methods have be established for the SPD task.
  (3) We propose U2P matching networks with a fine granularity to explore the matching between two sets of sentences, which outperform their C2P counterparts with a coarse granularity on the SPD task in our experiments.
  We hope the task and datasets will invite more research on detecting speaker profiles from conversational texts.

\section{Related Work}
    
  \paragraph{Speaker Profile in Text}
  Maintaining the consistency between a speaker and its utterances is an important issue in many NLP tasks. 
  A bunch of work investigates how to generate or select a dialogue response which is consistent with a specific speaker for building personalized chatbots.
  The prior descriptions of the speaker are usually presented by a persona which is composed of multiple profiles.
  \citet{li2016persona} proposed a persona-based neural conversation generation model to capture individual characteristics such as background information and speaking style.
  \citet{zhang2018personalizing} constructed a \texttt{Persona-Chat} dataset using an artificial data collection mechanism based on Mechanical Turk.
  As a result, neither dialogues nor personas can be fully representative of real user-agent interactions and the dataset coverage remains limited, containing a bit more than 1K different personas.
  To imitate the real-life scenarios, \citet{mazare2018training} constructed a persona-based dialogue dataset using conversations previously extracted from Reddit, where personas are collected with simple heuristic rules.
  \citet{gu2019dually} proposed a dually interactive matching network which adopted a dual matching architecture for presenting the personalities of dialogue agents in retrieval-based chatbots.
  \citet{hua2020learning,gu2020filtering,zhu2021content} attempt to learn to detect relevant contexts and user profiles in retrieval-based chatbots.
  \citet{gu2021partner} thoroughly explore the impact of utilizing personas that describe either self or partner speakers on the task of response selection in retrieval-based chatbots.

  \paragraph{Matching Tasks}
  Matching aims at searching for a best-matched one from a list of candidates.
  Determining the semantic matching degree or label between two pieces of text is a basic problem in many natural language understanding tasks.
  The existing matching or classification tasks can be generally categorized into two main categories that establish the relationship either (1) between a pair of sentences \cite{wang2013dataset,bowman2015large,williams2018broad}, or (2) between a set of sentences and a single sentence \cite{lowe2015ubuntu,lai2017natural,wu2017sequential,han2018fewrel}.
  We name them \emph{one-to-one} and \emph{many-to-one} matching in this paper.

  Different from the studies mention above, our proposed SPD task is a new \emph{many-to-many} semantic matching task, whose important characteristic is that both contexts and personas are composed of multiple sentences.
  This discourse-level matching is challenging since the informativeness of an utterance or a profile is dynamically changing, which requires models to filter out the redundant information automatically.
  Thus, we propose the utterance-to-profile (U2P) matching networks which are effective at filtering out less informative sentences and obtaining a more informative representation for the whole set.

\section{Dataset Creation}

  \begin{table}[t]
     \centering
     \setlength{\tabcolsep}{0.2mm}
     \begin{tabular}{cc|ccc}
      \toprule
                            &                                 & Train  & Valid  & Test   \\
      \hline
       \multirow{1}*{10@1}  & \# distractors (\emph{N})       &    1   &   9    &   9 \\
       \multirow{1}*{100@1} & \# distractors (\emph{N})       &    1   &   99   &   99    \\
      \hline
       \multicolumn{2}{c|}{\# matched context-persona pairs}  &  18K   &   2K   &   2K   \\
       \multicolumn{2}{c|}{Avg. \# utterances per context}    &  7.35  &  7.80  &  7.76  \\
       \multicolumn{2}{c|}{Avg. \# words per utterance}       &  11.67 &  11.94 &  11.79 \\
       \multicolumn{2}{c|}{Avg. \# profiles per persona}      &  4.50  &  4.49  &  4.50  \\
       \multicolumn{2}{c|}{Avg. \# words per profile}         &  7.32  &  7.82  &  7.56  \\
      \bottomrule
      \end{tabular}
      \caption{Statistics of the \texttt{PMPC}, where 10@1 and 100@1 correspond to using 9 and 99 distractors respectively in the validation and test sets.}
      \label{tab2}
  \end{table}
  
  A dataset named \texttt{Persona Match on Persona-Chat (PMPC)} is built for studying the SPD task. 
  Luckily, our data construction can be based on an existing \texttt{Persona-Chat} dataset \cite{zhang2018personalizing} which is a high-quality one made available at MILA and Facebook AI.
  Each dialogue in \texttt{Persona-Chat} was created by first assigning each human speaker a random persona and then asking them to chat in pair conditioned on the given personas.
  The personas in \texttt{Persona-Chat} are created by crowd-sourcing and modified by human.
  Profiles in a persona co-refer to each other, and each profile describes different properties of a coherent persona.
  Due to the natural dataset creating method, the conversational texts conditioned on the given personas can intuitively reflect characteristics of the speaker, which creates the natural matching relationship between the conversational texts and the persona of a speaker.
  Thus, we assume that \texttt{Persona-Chat} is suitable for studying the SPD task.
  Since each dialogue in \texttt{Persona-Chat} was performed between two speakers, we can consider one of them as human speaker and the other as intelligent agent, and then exchange with each other.
  In this way, each dialogue in \texttt{Persona-Chat} can provide two matched context-persona pairs.
  Two versions of personas, including original ones and revised ones were provided by \texttt{Persona-Chat}.
  The latter is constructed by rephrasing, generalizing, or specializing the original one.
  The revised version is adopted in this paper to make the SPD task more challenging.

  The task of SPD is defined as selecting a best-matched persona from a list of candidates according to the conversational texts of the speaker, as shown in Figure \ref{fig4}.
  The candidate set is composed of one correct persona and \emph{N} incorrect personas, which we call \textit{distractors}.
  The context is used as the input conversational texts for detecting speaker personas.
  The matched persona of the context is adopted to represent the correct persona in the candidate set.
  Besides, \emph{N} random persona distractors are chosen from the persona pool to form the complete candidate set.
  Table~\ref{tab2} presents the statistics of the \texttt{PMPC} dataset.
  Two configurations of \emph{N} (i.e., 9 and 99) are used to construct the validation and test sets.
  There is no overlap on contexts and personas among the training, validation and test sets in \texttt{Persona-Chat}.
  We follow this setting by constructing context-persona pairs in the respective sets.
  Thus, there is no overlap in \texttt{PMPC} as well.

\section{Models}
  We present the formal definition of SPD as follows.
  Given a dataset $\mathcal{D}$, an example of the dataset can be represented as $(c,p,y)$.
  Specifically, $c = \{u_1,u_2,...,u_{n_c}\}$ represents a context with $\{u_m\}_{m=1}^{n_c}$ as its utterances and $n_c$ as the utterance number.
  $p = \{p_1,p_2,...,p_{n_p}\}$ represents a persona with $\{p_n\}_{n=1}^{n_p}$ as its profiles and $n_p$ as the profile number similarly.
  $y \in \{0,1\}$ denotes a label. $y=1$ indicates that $c$ and $p$ are a matched pair; otherwise, $y=0$.
  Our goal is to learn a matching model $g(c,p)$ from $\mathcal{D}$.
  For any pair $(c,p)$, $g(c,p)$ measures the matching degree between $c$ and $p$.
  We learn $g(c,p)$ by minimizing its cross-entropy loss on $\mathcal{D}$.
  Let $\Theta$ denote the parameters of the matching model.
  The objective function $\mathcal{L}(\mathcal{D}, \Theta)$ of learning can be formulated as
    \begin{equation}
    \begin{aligned}
      \mathcal{L}(\mathcal{D}, \Theta) = - \sum_{(c,p,y)\in \mathcal{D}} [ & y log(g(c,p)) \\ + (1-y) & log(1-g(c,p)) ].
    \end{aligned}
    \end{equation}

  Sentence-encoding-based, cross-attention-based and pretraining-based frameworks are followed to build matching models.
  Under each framework, C2P and U2P matching networks are designed, adopting the \emph{one-to-one} and \emph{many-to-many} matching respectively.
  The details of these models are introduced in the following subsections.

  \subsection{Sentence-Encoding-Based Models}

    Under this framework, three types of sentence encoders, BOW, BiLSTM and Transformer, are employed.
    They share the same model architecture except the sentence encoder.
    First, BOW is employed to explore whether simple \emph{n}-gram overlap could solve this task easily, which can prove the importance of considering the long-term dependency in this task.
    Second, BiLSTM and Transformer are employed to discuss the impact of chronological (BiLSTM) or parallel (Transformer) encoding on this task.
    Due to the space limit, we omit the descriptions on basic BOW, BiLSTM or Transformer.
    Readers can refer to \citet{joulin2017bag}, \citet{hochreiter1997long} and \citet{vaswani2017attention} for more details.

    \begin{figure}[t]
      \centering
      \subfigure[C2P-BOW/BiLSTM/Transformer]{
      \includegraphics[width=7.5cm]{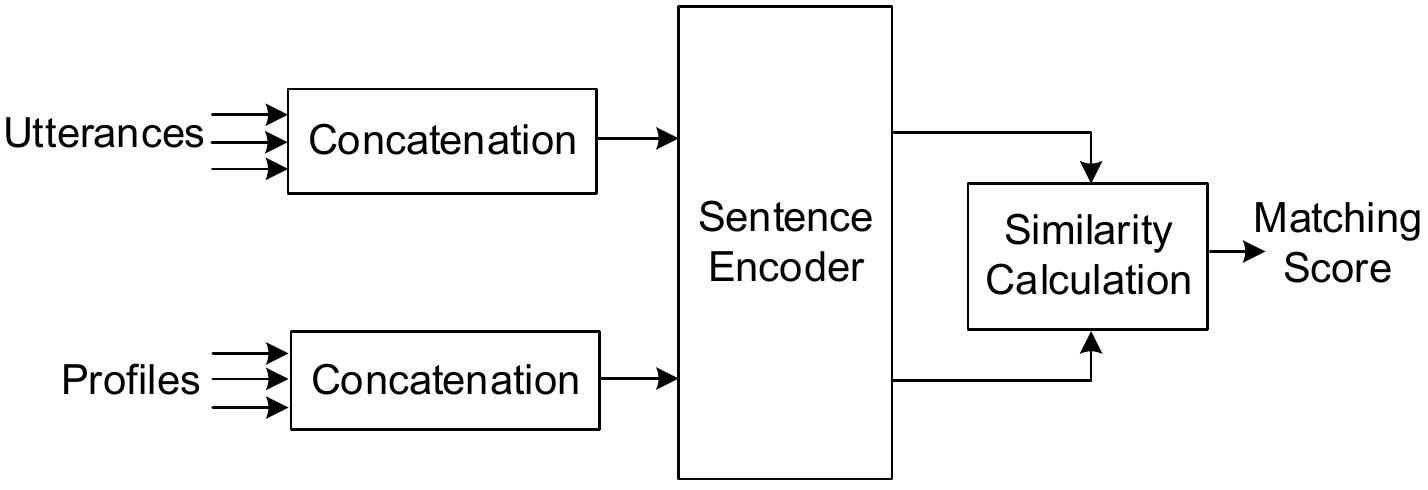}}
      \subfigure[U2P-BOW/BiLSTM/Transformer]{
      \includegraphics[width=7.5cm]{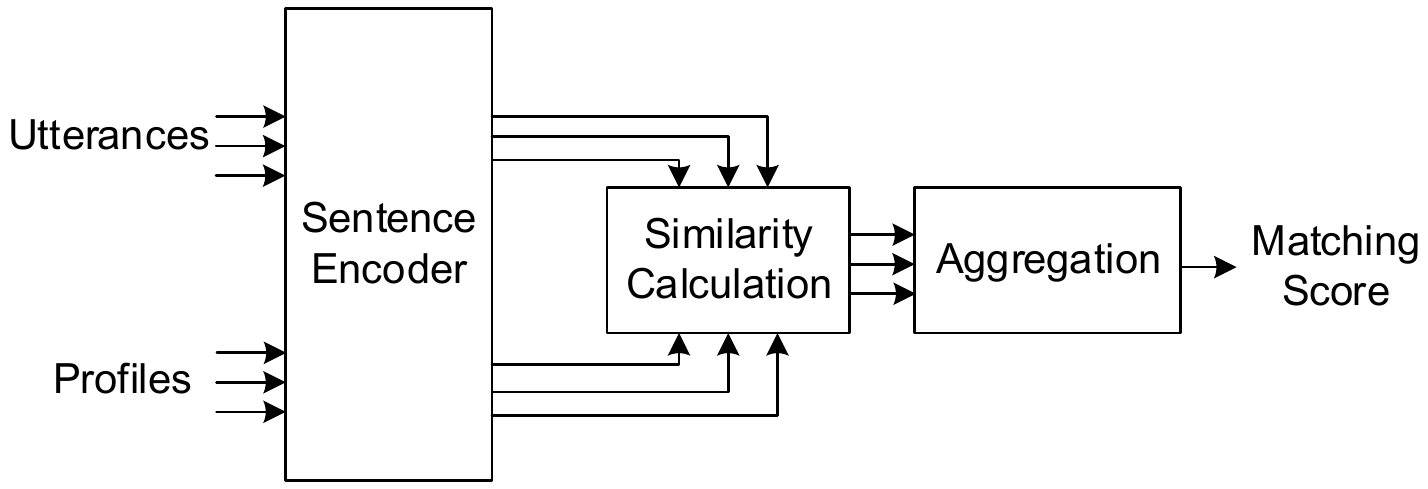}}
      \caption{Model architectures of (a) C2P-BOW/BiLSTM/Transformer and (b) U2P-BOW/BiLSTM/Transformer.}
      \label{fig1}
    \end{figure}

    \subsubsection{C2P-BOW/BiLSTM/Transformer}
      Figure~\ref{fig1}(a) presents the model architecture of C2P-BOW/BiLSTM/Transformer.
      Each utterance $u_m$ and each profile $p_n$ are first converted to their respective embedding matrices $\textbf{U}_m = \{\textbf{u}_{m,i}\}_{i=1}^{l_{u_m}}$ and $\textbf{P}_n = \{\textbf{p}_{n,j}\}_{j=1}^{l_{p_n}}$ by looking up a word embedding table, where $l_{u_m}$ and $l_{p_n}$ are the lengths of $u_m$ and $p_n$.
      Each $\textbf{u}_{m,i}$ or $\textbf{p}_{n,j}$ is an embedding vector.
      Then, the context representations $\textbf{C} = \{\textbf{c}_i\}_{i=1}^{l_c}$ with $l_c = \sum_{m=1}^{n_c} l_{u_m}$ are formed by concatenating the set of utterance representations $\{\textbf{U}_m\}_{m=1}^{n_c}$.
      The persona representations $\textbf{P} = \{\textbf{p}_j\}_{j=1}^{l_p}$ with $l_p = \sum_{n=1}^{n_p} l_{p_n}$ are formed similarly.
      Furthermore, the context representations $\textbf{C}$ and the persona representations $\textbf{P}$ are encoded by a shared sentence encoder, which is one of BOW, BiLSTM or Transformer.
      A max pooling operation is followed to derive the vectors $\bar{\textbf{c}} \in \mathbb{R}^d$ and $\bar{\textbf{p}} \in \mathbb{R}^d$ for representing the context and the persona.
      Finally, a similarity matrix \textbf{A} $\in \mathbb{R}^{d \times d}$ and a sigmoid function $\sigma$ are used to compute the final matching degree $g(c,p)$ as follows,
      \begin{align}
        g(c,p) = ~\sigma(~\bar{\textbf{c}}^\top \cdot \textbf{A} \cdot \bar{\textbf{p}}),
      \end{align}
      where $g(c,p)$ denotes the matching score for a $(c,p)$ pair.

    \subsubsection{U2P-BOW/BiLSTM/Transformer}
      Figure~\ref{fig1}(b) shows the model architecture of U2P-BOW/BiLSTM/Transformer.
      In these models, the word representations are constructed in the same way as their C2P counterparts.
      Instead of concatenating first, each utterance $\textbf{U}_m$ and each profile $\textbf{P}_n$ is encoded in parallel and separately by one of BOW, BiLSTM or Transformer encoder.
      Then, the same pooling operation is applied, and a set of utterance representations $\{\bar{\textbf{u}}_m\}_{m=1}^{n_c}$ and a set of profile representations $\{\bar{\textbf{p}}_n\}_{n=1}^{n_p}$ are derived.
      A similarity score $s_{mn}$ is computed for each utterance-profile pair as follows,
      \begin{equation}
        s_{mn} = \bar{\textbf{u}}_m^\top \cdot \textbf{A} \cdot \bar{\textbf{p}}_n.
      \end{equation}

      In order to obtain the matching score between the whole set of utterances and the whole set of profiles, additional aggregation operations are required.
      Here, we make an assumption that one utterance can only reflect one profile.
      Thus, for a given utterance, its matching score with the persona is defined as the maximum matching score between it and all profiles.
      If no profile provides positive scores for an utterance, this utterance is considered as uninformative and its matching score is set to 0.
      Finally, we accumulate the matching scores of all utterances and derive the final matching score $g(c,p)$.
      Mathematically, we have
      \begin{align}
        s_m    &= \max \{ \max \limits_{n} \ s_{mn} , 0 \}, \label{equ1} \\
        s      &= \sum \limits_{m=1}^{n_c} s_m, \label{equ2} \\
        g(c,p) &= \sigma(s), \label{equ3}
      \end{align}
      More verification about the aggregation methods and our assumption will be discussed in Section of Experiments.

  \subsection{Cross-Attention-Based Models}

    ESIM \cite{chen2017enhanced} is adopted as the basis of models under this framework.

    \begin{figure}[t]
      \centering
      \subfigure[C2P-ESIM]{
      \includegraphics[width=3.7cm]{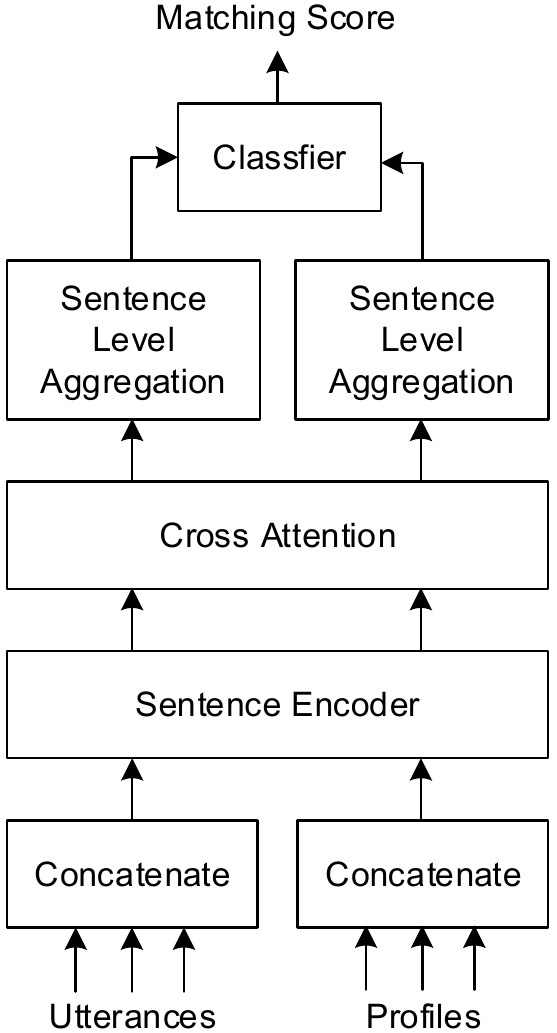}}
      \subfigure[U2P-ESIM]{
      \includegraphics[width=3.7cm]{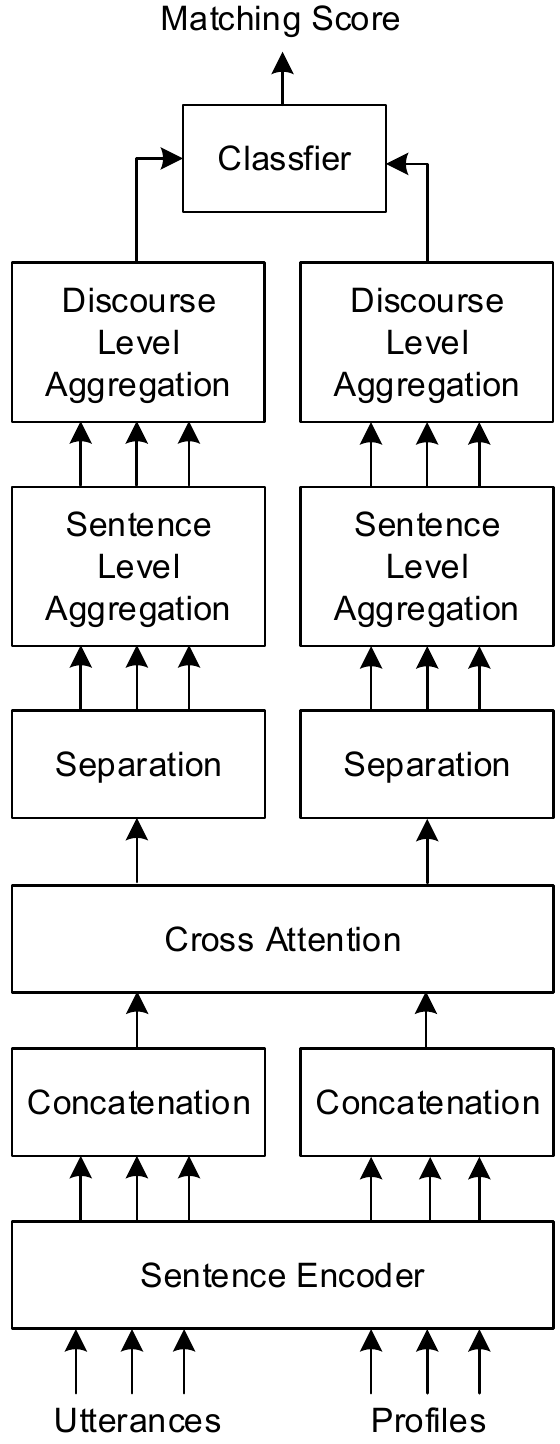}}
      \caption{Model architectures of (a) C2P-ESIM and (b) U2P-ESIM.}
      \label{fig2}
    \end{figure}

    \subsubsection{C2P-ESIM}
      
      Figure~\ref{fig2}(a) presents the model architecture of C2P-ESIM.
      It shares the same word representation, sentence concatenation and sentence encoder as those in C2P-BiLSTM to derive the encoded context and persona representations
      $\bar{\textbf{C}} = \{\bar{\textbf{c}}_i\}_{i=1}^{l_c}$ with $l_c = \sum_{m=1}^{n_c} l_{u_m}$ and $\bar{\textbf{P}} = \{\bar{\textbf{p}}_j\}_{j=1}^{l_p}$ with $l_p = \sum_{n=1}^{n_p} l_{p_n}$.
      Then, an attention-based alignment is performed to obtain $\textbf{C}^{mat}$ and $\textbf{P}^{mat}$.
      Furthermore, sentence-level aggregation operations are performed to obtain the context and persona embedding vectors $\textbf{c}^{agr}$ and $\textbf{p}^{agr}$.
      Finally, the matching feature vector is formed by concatenating $\textbf{c}^{agr}$ and $\textbf{p}^{agr}$, and is sent into a multi-layer perceptron (MLP) classifier for calculating $g(c,p)$.

    \subsubsection{U2P-ESIM}
      Figure~\ref{fig2}(b) presents the model architecture of U2P-ESIM.
      Different from C2P-ESIM which encodes long sequences after concatenating context utterances and persona profiles, U2P-ESIM first encodes each sentence in parallel and separately.
      It first encodes each utterance and profile separately to derive $\{\bar{\textbf{U}}_m\}_{m=1}^{n_c}$ and $\{\bar{\textbf{P}}_n\}_{n=1}^{n_p}$.
      To apply the attention-based alignment between a $(c,p)$ pair, $\{\bar{\textbf{U}}_m\}_{m=1}^{n_c}$ are concatenated to form the encoded context representations $\bar{\textbf{C}} = \{\bar{\textbf{c}}_i\}_{i=1}^{l_c}$ with $l_c = \sum_{m=1}^{n_c} l_{u_m}$.
      The encoded persona representations $\bar{\textbf{P}} = \{\bar{\textbf{p}}_j\}_{j=1}^{l_p}$ are formed similarly with $l_p = \sum_{n=1}^{n_p} l_{p_n}$
      The same cross attention used in C2P-ESIM is performed between $\bar{\textbf{C}}$ and $\bar{\textbf{P}}$ to obtain $\textbf{C}^{mat}$ and $\textbf{P}^{mat}$, which are further converted back into separate utterances $\{\textbf{U}_m^{mat}\}_{m=1}^{n_c}$ and profiles $\{\textbf{P}_n^{mat}\}_{n=1}^{n_p}$ according to the lengths of individual sentences.
      Then, the same sentence-level aggregation in C2P-ESIM is applied to process each $\textbf{U}_m^{mat}$ and $\textbf{P}_n^{mat}$ to derive a set of utterance embeddings $\{\textbf{u}_{m}^{agr}\}_{m=1}^{n_c}$ and a set of profile embeddings $\{\textbf{p}_{n}^{agr}\}_{n=1}^{n_p}$.
      An additional discourse-level aggregation is performed over $\{\textbf{u}_{m}^{agr}\}_{m=1}^{n_c}$ and $\{\textbf{p}_{n}^{agr}\}_{n=1}^{n_p}$ to get the embedding vectors $\textbf{c}^{agr}$ for the whole context and $\textbf{p}^{agr}$ for the whole persona.
      Different aggregation strategies are designed for them.
      As the context utterances are extracted from a conversation and their chronological relationships are maintained, $\{\textbf{u}_{m}^{agr}\}_{m=1}^{n_c}$ are sent into another BiLSTM and aggregated by a pooling operation. 
      On the other hand, the profiles in a persona have no chronological orders.
      Thus, an attention-based aggregation strategy is employed which first projects each profile vector to a scalar and then computes the attention weight for each profile.
      The aggregated persona representation $\textbf{p}^{agr}$ is calculated as the weighted sum of $\{\textbf{p}_{n}^{agr}\}_{n=1}^{n_p}$.
      Finally, the matching feature is also the concatenation of $\textbf{c}^{agr}$ and $\textbf{p}^{agr}$, which is sent into an MLP classifier to return the matching degree.

    \begin{table*}[t]
     \centering
     \begin{tabular}{l|cc|cc}
      \toprule
       Model                      & $\textbf{R}_{10}@1$ & $\textbf{MRR}_{10}$ & $\textbf{R}_{100}@1$ & $\textbf{MRR}_{100}$ \\
      \hline
       C2P-BOW                    & 34.7 $\pm$ 1.2 & 54.4 $\pm$ 0.9 &  8.9 $\pm$ 0.5 & 19.5 $\pm$ 0.5  \\
       U2P-BOW                    & 46.5 $\pm$ 1.7 & 63.3 $\pm$ 1.3 & 16.9 $\pm$ 1.2 & 28.5 $\pm$ 1.2  \\
      \hline
       C2P-BiLSTM                 & 38.3 $\pm$ 1.2 & 57.7 $\pm$ 0.9 &  8.1 $\pm$ 0.8 & 19.2 $\pm$ 0.9  \\
       U2P-BiLSTM                 & 57.4 $\pm$ 1.4 & 71.0 $\pm$ 1.4 & 24.0 $\pm$ 1.6 & 37.5 $\pm$ 1.6  \\
      \hline
       C2P-Transformer            & 49.6 $\pm$ 3.7 & 65.3 $\pm$ 2.5 & 19.0 $\pm$ 1.5 & 30.5 $\pm$ 1.1 \\
       U2P-Transformer            & 56.2 $\pm$ 1.5 & 70.6 $\pm$ 1.1 & 22.9 $\pm$ 1.3 & 36.0 $\pm$ 1.3 \\
      \hline
       C2P-ESIM                   & 80.7 $\pm$ 0.5 & 87.7 $\pm$ 0.4 & 50.7 $\pm$ 1.4 & 62.8 $\pm$ 0.7 \\
       U2P-ESIM                   & 81.6 $\pm$ 1.0 & 88.4 $\pm$ 0.6 & 54.5 $\pm$ 1.3 & 66.6 $\pm$ 0.7 \\
      \hline
       C2P-BERT                   & 87.4 $\pm$ 0.7 & 91.8 $\pm$ 0.4 & 64.7 $\pm$ 1.5 & 75.4 $\pm$ 0.8 \\
       U2P-BERT                   & 90.4 $\pm$ 0.5 & 94.3 $\pm$ 0.2 & 79.1 $\pm$ 0.9 & 83.2 $\pm$ 0.5 \\
      \bottomrule
      \end{tabular}
      \caption{Evaluation results (mean $\pm$ standard deviation) (\%) of different models on the test set of \texttt{PMPC} in which each test sample had either 9 or 99 distractors.}
      \label{tab3}
    \end{table*}

  \subsection{Pretraining-Based Models}
  
    BERT \cite{devlin2019bert} is adopted as the basis of models under this framework.
    
    \begin{figure}[t]
      \centering
      \subfigure[C2P-BERT]{
      \includegraphics[width=7.5cm]{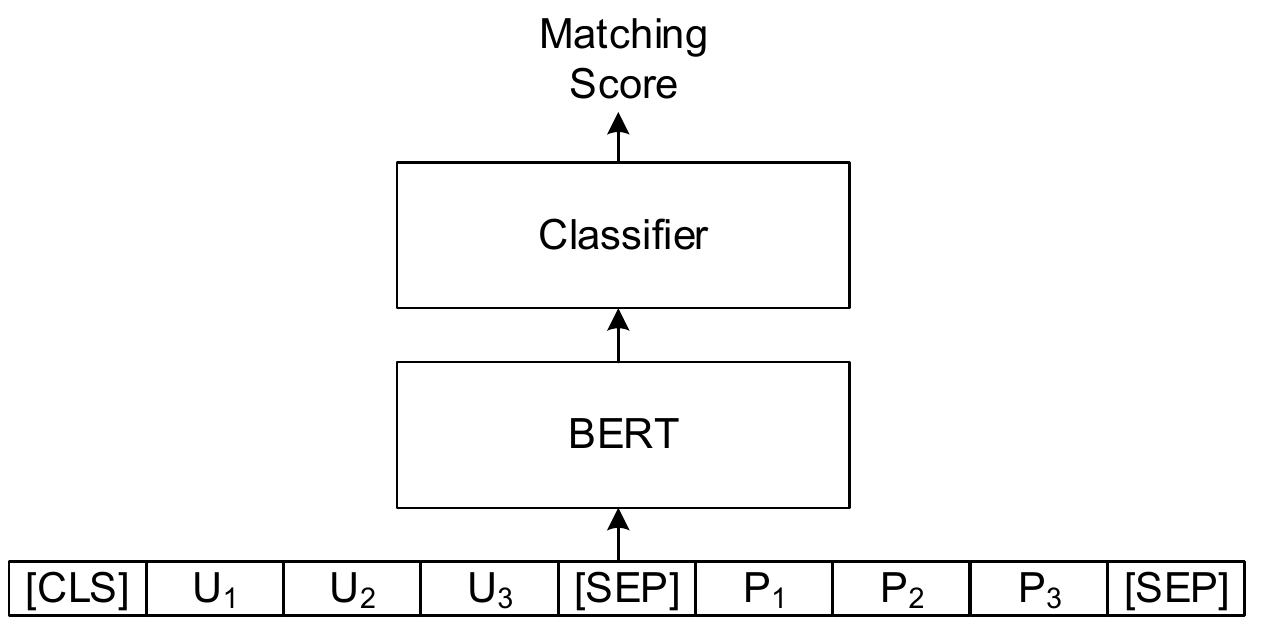}}
      \subfigure[U2P-BERT]{
      \includegraphics[width=4.6cm]{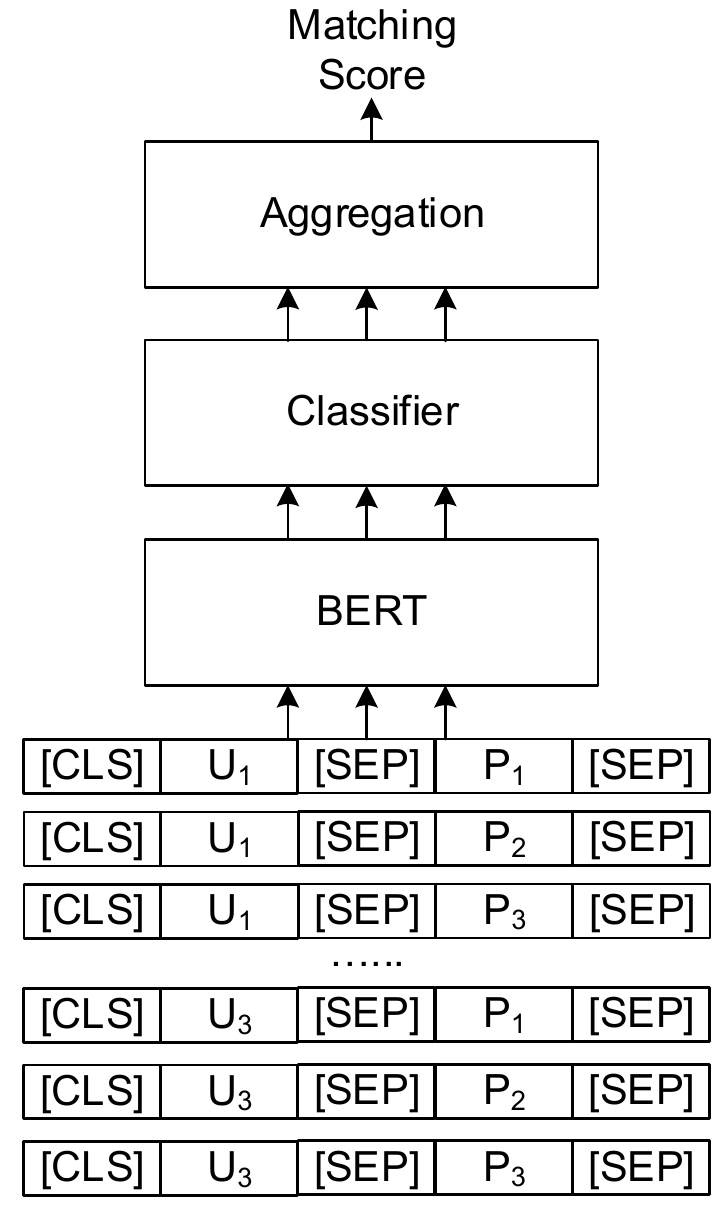}}
      \caption{Model architectures of (a) C2P-BERT and (b) U2P-BERT.}
      \label{fig5}
    \end{figure}

    \subsubsection{C2P-BERT}
      Figure~\ref{fig5}(a) presents the model architecture of C2P-BERT.
      In BERT, sentence A and sentence B are concatenated with a \texttt{[SEP]} token, which is then sent into BERT and classified through the representation of the \texttt{[CLS]} token.
      Here, utterances are concatenated to form the sentence A, and profiles are concatenated to form the sentence B.
      The token type schema of C2P-BERT in our current manuscript strictly follows the original BERT to provide fair comparisons. 
      Interactions are performed between the whole context and the whole persona at a coarse granularity through stacked Transformer blocks.
      Finally, the representation of the \texttt{[CLS]} token is sent into an MLP classifier to return the matching degree $g(c,p)$.

    \subsubsection{U2P-BERT}
      Figure~\ref{fig5}(b) presents the model architecture of U2P-BERT.
      In order to perform the matching at a finer granularity, interactions and matching are performed between each utterance and each profile.
      In detail, for a specific utterance $u_m$ which is used to form the sentence A, it is concatenated with each profile $p_n$ which is used to form the sentence B respectively.
      The same concatenation operation is performed for each utterance.
      In this way, we can derive several sets of concatenated sequences.
      These concatenated sequences are sent into BERT for encoding in parallel.
      Their output \texttt{[CLS]} representations are sent into an MLP classifier to return several sets of similarity score $s_{mn}$, which denotes the matching degree between $u_m$ and $p_n$.
      Finally, additional aggregation operations as introduced by Eq.~(\ref{equ1}), Eq.~(\ref{equ2}) and Eq.~(\ref{equ3}) are applied to get $g(c,p)$ denoting the final matching degree between the whole set of utterances and the whole set of profiles.

\section{Experiments}

  \subsection{Evaluation Metrics}
    We adopted the evaluation metrics popularly used by other text retrieval tasks \cite{lowe2015ubuntu,wu2017sequential,zhang2018personalizing}.
    Each model was tasked with selecting the $k$ best-matched ones from $n$ available candidates, and we calculated the recall of the true positive one among the $k$ selected candidates, denoted as $\textbf{R}_n@k$. 
    In addition, mean reciprocal rank, denoted as $\textbf{MRR}_{n}$, was also calculated, which was the average of reciprocal ranks of retrieval results among $n$ available candidates.
  
  \subsection{Training Details}
    For BOW, BiLSTM, Transformer and ESIM, the word representations were 300-dimensional GloVe embeddings \cite{pennington2014glove} and were not updated during training.
    All hidden states of the LSTM had 200 dimensions.
    The MLP at the prediction layer had 256 hidden units with ReLU \cite{nair2010rectified} activation.
    Dropout \cite{srivastava2014dropout} with a rate of 0.2 was applied to the word embeddings and all hidden layers.
    The maximum utterance length, maximum number of utterances in a context, maximum profile length and maximum number of profiles in a persona were set to 20, 8, 15 and 5 respectively for \texttt{PMPC}.
    Zeros were padded if the number of utterances in a context and the number of profiles in a persona were less than the limits.
    Otherwise, we kept the last 8 utterances in the context and the last 5 profiles in the persona.
    The Adam method \cite{kingma2014adam} was employed for optimization. 
    The learning rate was initialized as 1e-3 and was exponentially decayed by 0.96 every 5000 steps.
    The validation set was utilized to select the best model for testing.
    For BERT, we adopted the base version whose number of Transformer block is set to 12, hidden size is set to 768 and number of self-attention heads is set to 12.
    The learning rate was initialized as 2e-5.

    All codes were implemented under the TensorFlow framework \cite{abadi2016tensorflow} and are published along with the dataset to help replicate our results.~\footnote{https://github.com/JasonForJoy/SPD}

  \subsection{Experimental Results}

    Table~\ref{tab3} presents the evaluation results of different models on the test set of \texttt{PMPC}. 
    Each model was trained for 10 times with identical architectures and different random initializations. 
    It can be noticed that all U2P models achieved better performance than their C2P counterparts on all metrics.
    Specifically, U2P-BOW, U2P-BiLSTM, U2P-Transformer, U2P-ESIM and U2P-BERT outperformed their C2P counterparts by margins of 11.8\%, 19.1\%, 6.6\%, 0.9\% and 3.0\% respectively in terms of $\textbf{R}_{10}@1$, and by margins of 8.0\%, 15.9\%, 3.9\%, 3.8\% and 14.4\% respectively in terms of $\textbf{R}_{100}@1$.
    These results demonstrate the effectiveness of treating both contexts and personas as sets of multiple sequences and conducting \emph{many-to-many} matching for this task. 
    Comparing BOW, BiLSTM and Transformer, it can be seen that it is actually necessary to model the long-term dependency for this task, as a simple \emph{n}-gram overlap based method could not perform competitively with models considering the long-term dependency.
    Comparing BiLSTM with Transformer at both coarse and fine granularities, BiLSTM achieved larger improvement than Transformer after employing the U2P framework, which shows its effectiveness especially for models with chronological encoders.
    Comparing C2P-BERT with U2P-BERT, although C2P-BERT has been the state-of-the-art model for capturing the matching information, U2P-BERT could still provide additional improvement.
    U2P-BERT achieves only 79.1\% $\textbf{R}_{100}@1$, which shows that this task is difficult and there is a lot of room for further improvement.

  \subsection{Analysis}

    \begin{table}[t]
     \centering
     \setlength{\tabcolsep}{1.5mm}{
     \begin{tabular}{ccc}
      \toprule
       Aggregation Strategy           & $\textbf{MRR}_{10}$ & $\textbf{R}_{10}@1 $ \\
      \midrule
       Ps-\emph{Max} \& Us-\emph{Sum} & 71.0 $\pm$ 1.4      & 57.4 $\pm$ 1.4  \\
       Ps-\emph{Max} \& Us-\emph{Max} & 70.0 $\pm$ 1.3      & 53.7 $\pm$ 1.9  \\
       Ps-\emph{Sum} \& Us-\emph{Max} & 57.3 $\pm$ 0.8      & 37.1 $\pm$ 1.0  \\
       Ps-\emph{Sum} \& Us-\emph{Sum} & 67.5 $\pm$ 0.7      & 51.5 $\pm$ 1.1  \\
      \bottomrule
      \end{tabular}}
      \caption{Evaluation results (\%) of U2P-BiLSTM models with different aggregation strategies on the test set of \texttt{PMPC} (\emph{N}~=~9).
     Ps denotes Profiles and Us denotes Utterances.
     \emph{Max} and \emph{Sum} denote the aggregation operation used in Eq.~(\ref{equ1}) and Eq.~(\ref{equ2}).}
      \label{tab5}
    \end{table}

    \paragraph{Aggregation Method}
    One characteristic of U2P matching networks is that the discourse-level aggregation is necessary to assemble the matching scores for each utterance-profile pair.
    As introduced by Eq.~(\ref{equ1}) and Eq.~(\ref{equ2}), the aggregation operation employs \emph{Max} over the set of profiles to select the best-matched profile for each utterance, and then employs \emph{Sum} over the set of utterances to accumulate the matching scores for all utterances.
    In order to verify the effectiveness of this strategy, we replaced it with other settings and evaluated them on the test set as shown in Table~\ref{tab5}.
    By comparing the first row with the fourth row, as well as the second row with the third row, we can see that \emph{Max} achieved better performance of aggregating profiles than \emph{Sum}, which supports our assumption that one utterance may reflect only one profile.
    Then, by comparing the first row with the second row, as well as the third row with the fourth row, we can see that \emph{Sum} achieved better performance of aggregating utterances than \emph{Max}, which indicates that multiple utterances should be considered when deriving the matching score for a context-persona pair.

    \begin{table}[t]
     \centering
     \setlength{\tabcolsep}{1.6mm}{
     \begin{tabular}{c|cccccccc}
      \toprule
                 &   U1  &   U2  &  U3   &  U4   &  U5   &  U6     \\
      \hline
       P1        & -0.07 & -0.35 & -0.22 & -0.70 & -1.05 & -0.19   \\
       P2        & -0.16 &  0.90 &  0.72 & -0.20 &  0.38 & -0.34   \\
       P3        &  0.83 &  1.14 &  1.00 & -0.48 &  0.05 & -0.10   \\
       P4        & -0.92 & -1.17 & -0.89 & -0.64 & -2.21 & -0.09   \\
      \hline
       $s_{m}$   &  0.83 &  1.14 &  1.00 &  0.0  &  0.38 &  0.0     \\
      \bottomrule
      \end{tabular}}
      \caption{Utterance-profile similarity scores for the matched context-persona pair shown in Figure~\ref{fig4}. Here, U$m$ and P$n$ denote the $m$-th utterance and the $n$-th profile respectively.}
      \label{tab7}
    \end{table}

    \paragraph{Case Study}
    We further illustrate the interpretability of the aggregation operation in U2P models by conducting a case study as shown in Table~\ref{tab7}, which contains the utterance-profile similarity scores for the matched context-persona pair shown in Figure~\ref{fig4}.
    First, we can see that these matching scores computed by U2P models are reasonable.
    For example, U1 (\emph{I actually just got back from a hike}) and U5 (\emph{I am single mom}) achieved high similarity scores with P3 (\emph{I enjoy nature walks}) and P2 (\emph{I am raising sons all on my own}) respectively.
    Second, the $s_m$ values calculated by Eq.~(\ref{equ1}) indicate that some utterances may be uninformative and the model filtered them out automatically when calculating the final score for a context-persona pair.
    Scoring each utterance-profile pair at a fine granularity is effective to select the best-matched profile for each utterance and filter out uninformative utterances at the same time.

    \begin{table}[t]
     \centering
     \begin{tabular}{lcc}
      \toprule
      Model                & Time (s)   & Parameters    \\
      \midrule
      C2P-BOW              & 7.1        & 90k    \\
      U2P-BOW              & 8.6        & 90k    \\
      \midrule
      C2P-BiLSTM           & 17.1       & 962K    \\
      U2P-BiLSTM           & 12.2       & 962K    \\
      \midrule
      C2P-Transformer      & 8.3        & 271K    \\
      U2P-Transformer      & 10.3       & 271K    \\
      \midrule
      C2P-ESIM             & 36.7       & 4.1M    \\
      U2P-ESIM             & 22.4       & 5.7M    \\
      \midrule
      C2P-BERT             & 121.3      & 110M    \\
      U2P-BERT             & 742.8      & 110M    \\
      \bottomrule
      \end{tabular}
      \caption{The inference time over the validation set of \texttt{PMPC} whose configuration of \emph{N} was 9 using different models, together with their numbers of parameters.}
      \label{tab4}
    \end{table}

    \paragraph{Complexity}
    We analysed the time and space complexity difference between the C2P and U2P models. 
    
    In order to explore the efficiency difference between C2P and U2P models, we analysed their time complexity by comparing their run-time computation speed.
    We recorded the inference time over the validation set of \texttt{PMPC} whose configuration of \emph{N} was 9 using a GeForce GTX 1080 Ti GPU.
    The inference time of each model is measured on a batch basis, instead of on a per-instance basis. 
    The results are shown in Table~\ref{tab4}.
    As we can see that, U2P-BOW was not as efficient as C2P-BOW.
    So did U2P-Transformer and C2P-Transformer.
    However, U2P-BiLSTM and U2P-ESIM were faster than their C2P counterparts by 28.7\% and 39.0\% respectively, although the U2P models relied on additional aggregation operations.
    The reason may be that the parallel encoding of multiple sequences in U2P networks can improve the efficiency of RNN-based sentence encoders but can not benefit the BOW-based or Transformer-based ones.
    U2P-BERT took more time than C2P-BERT as the calculation of former is an order of magnitude higher than the latter.
    However, the empirical improvement of U2P-BERT far outweighed the increased computation cost.

    The number of parameters of these C2P and U2P models was used to evaluate the space complexity of different models.
    The results are shown in Table~\ref{tab4}.
    We can see that C2P-BOW and U2P-BOW contained the same number of parameters.
    So did C2P-BiLSTM, C2P-Transformer and C2P-BERT with their U2P counterparts.
    The reason is that the additional aggregation operations in these U2P models consume only the calculation of \emph{Max} or \emph{Sum} function, while do not require additional parameters.
    U2P-ESIM adopted an additional BiLSTM for discourse-level aggregation, and thus contained more parameters than C2P-ESIM.
    
    These results show that the U2P models outperform their C2P counterparts significantly with comparable time and space complexity.

\section{Conclusion}
  In this paper, we propose the task of Speaker Persona Detection (SPD) and build a \texttt{PMPC} dataset for studying this task.
  In SPD, the matching from conversational contexts to persona candidates is established between two sets of sentences and requires a new \emph{many-to-many} matching framework. 
  Thus, this paper proposes the utterance-to-profile (U2P) matching networks, which treat both contexts and personas as sets of multiple sequences.
  Results show that the proposed U2P matching networks outperform their context-to-persona (C2P) counterparts, which treat both contexts and personas as single sequences.
  In the future, we will explore better aggregation methods and investigate model structures for \emph{many-to-many} matching.


\section*{Ethics}
  The \texttt{PMPC} dataset is transformed from the existing \texttt{Persona-Chat} dataset, personas and conversations of which are constructed by crowd-sourcing. We quote the description of the \texttt{Persona-Chat} dataset \cite{zhang2018personalizing} as follows: “We asked the crowdsourced workers to create a character (persona) description using 5 sentences. Because the personas are not the real profiles of the Turkers, the dataset does not contain personal information (and they are told specifically not to use any).” Furthermore, the personas and conversations are based on typical topics of daily life that speakers can bring up in conversation, without involving information such as person's attributes, race, etc. Thus, there are no ethical concerns in the \texttt{Persona-Chat} dataset and the \texttt{PMPC} dataset used in this paper.

\section*{Acknowledgements}
  We thank the anonymous reviewers for their valuable comments.

\bibliography{emnlp2021}
\bibliographystyle{acl_natbib}


\end{document}